\documentclass[conference]{IEEEtran}
\IEEEoverridecommandlockouts
\usepackage{cite}
\usepackage{amsmath,amssymb,amsfonts}
\usepackage{algorithmic}
\usepackage{graphicx}
\usepackage{textcomp}
\usepackage{xcolor}
\usepackage{booktabs}
\usepackage{multirow}
\def\BibTeX{{\rm B\kern-.05em{\sc i\kern-.025em b}\kern-.08em
    T\kern-.1667em\lower.7ex\hbox{E}\kern-.125emX}}
\begin{document}

\title{ A Ternary Bi-Directional LSTM Classification for Brain Activation Pattern Recognition Using fNIRS
{\footnotesize }
\thanks{}
}

\author{\IEEEauthorblockN{Sajila D. Wickramaratne}
\IEEEauthorblockA{\textit{Electrical and Computer Engineering} \\
\textit{University of New Hampshire}\\
Durham NH, United States \\
sdw1014@wildcats.unh.edu}

\and
\IEEEauthorblockN{MD Shaad Mahmud}
\IEEEauthorblockA{\textit{Electrical and Computer Engineering} \\
\textit{University of New Hampshire}\\
Durham NH, United States \\
mdshaad.mahmud@unh.edu}

}

\maketitle

\begin{abstract}
Functional near-infrared spectroscopy (fNIRS) is a non-invasive, low-cost method used to study the brain's blood flow pattern. Such patterns can enable us to classify performed by a subject. In recent research, most classification systems use traditional machine learning algorithms for the classification of tasks. These methods, which are easier to implement, usually suffer from low accuracy. Further, a  complex pre-processing phase is required for data preparation before implementing traditional machine learning methods. The proposed system uses a Bi-Directional LSTM based deep learning architecture for task classification, including mental arithmetic, motor imagery, and idle state using fNIRS data. Further, this system will require less pre-processing than the traditional approach, saving time and computational resources while obtaining an accuracy of 81.48\%, which is considerably higher than the accuracy obtained using conventional machine learning algorithms for the same data set. 
\end{abstract}

\begin{IEEEkeywords}
fNIRS, deep learning, LSTM, mental arithmetic
\end{IEEEkeywords}

\section{Introduction}

Functional near-infrared spectroscopy (fNIRS) is a  non-invasive and cost-effective neuroimaging technique. It utilizes near-infrared (NIR) light to measure the blood flow in the cortical regions of the brain\cite{naseer2015fnirs}.  The brain needs oxygen to perform tasks.  This oxygen is supplied through the blood. Hence, during brain activity, the demand for oxygenated blood increases, increasing the oxyhemoglobin(HbO2) level. The head layers, such as skin, skull, and lipid layers, are nearly transparent to the NIR light. The most absorbers of this light are HbO2 and deoxyhemoglobin (Hb) in the blood. Therefore, fNIRS uses two wavelengths in the NIR range to measure the concentration level of HbO2 and Hb by using the Beer-Lambert Law\cite{sassaroli2004comment}. 

Brain-computer interfaces (BCIs) have become a novel mode of communication for individuals who have lost the ability for voluntary movements \cite{blankertz2016berlin}\cite{abdalmalak2020assessing}.  BCIs can use a variety of neural signal recording methods, such as electroencephalography (EEG), magnetoencephalography, functional magnetic resonance imaging, near-infrared spectroscopy (NIRS), and electrocorticography \cite{abiri2019comprehensive}\cite{shin2018ternary}. Since each brain-imaging modality has its pros and cons, combining two or more neural signal recording modalities, which is generally referred to as a hybrid BCI, might enhance BCI's overall performance. Several hybrid BCI studies have demonstrated the effectiveness of the combinatory use of different modalities or paradigms\cite{hong2017hybrid}. Some examples include the steady-state visually evoked potential for EEG-BCI, hybrid EEG-electrooculogram (EOG) BCI, and hybrid EEG-NIRS BCI \cite{li2019advances}.

Although EEG can record brain activity from both frontal and central areas, NIRS find difficulty around the central area\cite{shin2018ternary}. This difficulty arises due to the attenuation of light intensity by the hair; hence without applying a time-consuming hair preparation process or using a particular type of optodes, it is difficult to obtain accurate NIRS data from the central area. Therefore in this dataset, NIRS signals were recorded only from the frontal area to improve the practicality. 

 In this paper, we propose the use of a multi-class fNIRS based system that classifies three brain activation patterns recorded during motor imagery (MI), mental arithmetic (MA), and idle state (IS) with a deep learning algorithm. MI has been the preferred mental task for EEG based BCI interfaces, while MA has been frequently used as a prevalent BCI task for NIRS based BCI interfaces. Brain activation response for MI tasks can be measured around the central area. In contrast, the response obtained by MA can be measured primarily in the forehead covering the prefrontal cortex (PFC)\cite{shin2018ternary}. The complex nature of the biological signals enables them to take advantage of deep learning networks' capabilities for classification systems. Although traditional machine learning methods are used frequently for fNIRS based task classification, in recent studies, deep learning classifiers have obtained satisfactory results\cite{tanveer2019enhanced} \cite{chiarelli2018deep}. The paper compares the different models that use various forms of data from the raw format, local features, and global features. The proposed system has a classification accuracy of 81.48\%, an improvement from the original classifier, which used only fNIRS data.

\section{Method}
For this paper, a deep learning model using LSTM was developed for ternary classification. This data was used to compare models that use different feature sets and even raw data. Deep Learning classifiers are powerful tools that can handle data with complex nonlinear relationships. The classification accuracy into varying levels of data preparation can help assess how much data pre-processing is required for the deep learning classifiers.

\subsection{System Overview}
\begin{figure*}[h]
        \centering
        \includegraphics[scale=0.6]{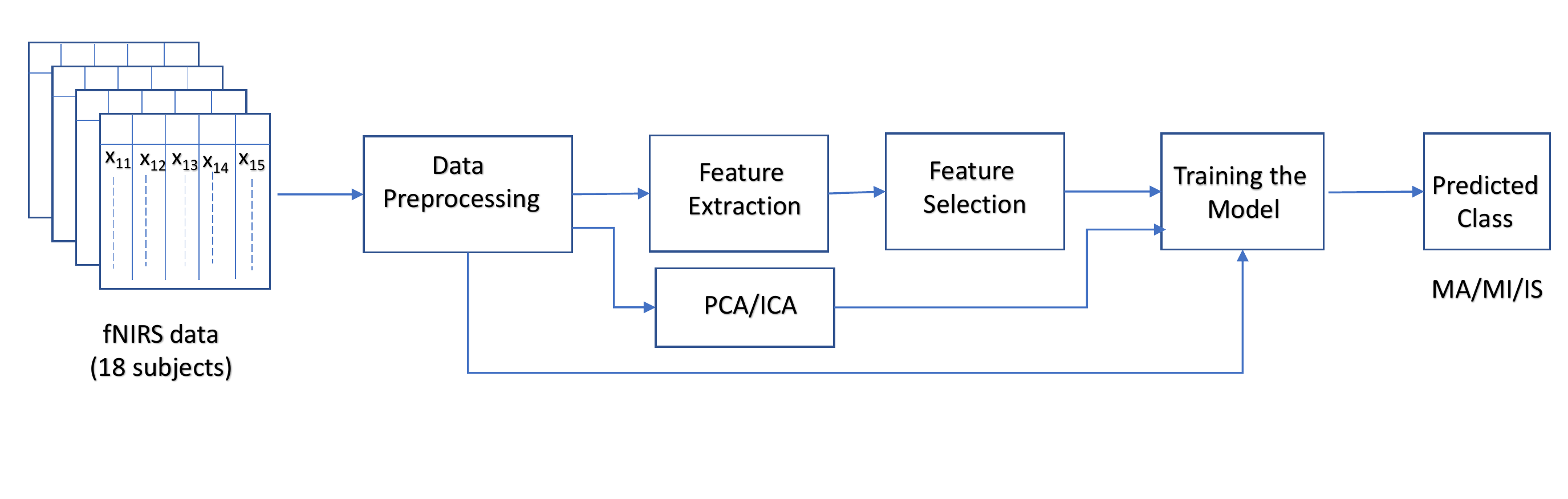}
        \caption{System Overview}
        \label{fig:sys_overview}
    \end{figure*}

The fNIRS data were analyzed in several different ways. The Fig.\ref{fig:sys_overview} illustrates the Overview of the complete systems. The individual components of the system will be described in detail in the later sections. The acquired data is initially pre-processed. This pre-processed data is directly fed into a deep learning network trained on only raw channel data. Another approach was to use dimension reduction techniques to pre-processed data and be fed to the deep learning network. The final approach was to extract features from the pre-processed data and use a feature selection mechanism to reduce dimensionality and train the neural network. After this phase, the data is fed to the model to be prepared. Finally, the result is obtained with the task being classified into MA, MI, or IS.

\subsection{Data}
The data was obtained from the study conducted by Shin et al. \cite{shin2018ternary}. The data consists of subjects performing three tasks: mental arithmetic, Motor Imagery, and Idle state. The subjects' demographics consisted of 10 men and eight women of  23.8 $\pm$ 2.5 years of age. NIRS data were collected using a portable NIRS system at a sampling rate of 13.3 Hz with 16 NIRS channels. Six sources and six detectors were placed on the forehead over the PFC. Each of these channels consisted of a source and detector pair set 30 mm away from each other.

The experimental procedure is fully explained in the paper by Shin et al. .\cite{shin2018ternary}. Each subject performs 90 tasks corresponding to the said tasks. The tasks are chosen randomly. Each trial follows the pattern of the introduction period, the selected task, and the resting period. 

During the introduction period, the participant will be informed which of the tasks will be performed by the participant.   For the right-hand MI task, the subjects performed an intricate finger-tapping pattern at approximately 2 Hz.  For the MA task, the subjects had to continuously subtract a one-digit number (between 6 and 9) from a former calculation as fast as possible.  For the IS task, the participants will stay relaxed without performing any specific mental imagery task. The subjects performed each of the tasks 30 times( 90 in total) in random order.

\subsubsection{Data Visualization}
The first step was to visualize the tasks in both the time domain and frequency domain. The figures show how the three tasks are represented in both the time and frequency domain. The visualization was done as the whole epoch, as well as just concentrating on the task. This visualization was essential before determining the features that should be extracted. Especially the frequency bands that should be considered when extracting frequency domain features were decided from the visualization. 

\begin{figure}[h]
        \centering
        \includegraphics[scale=0.25]{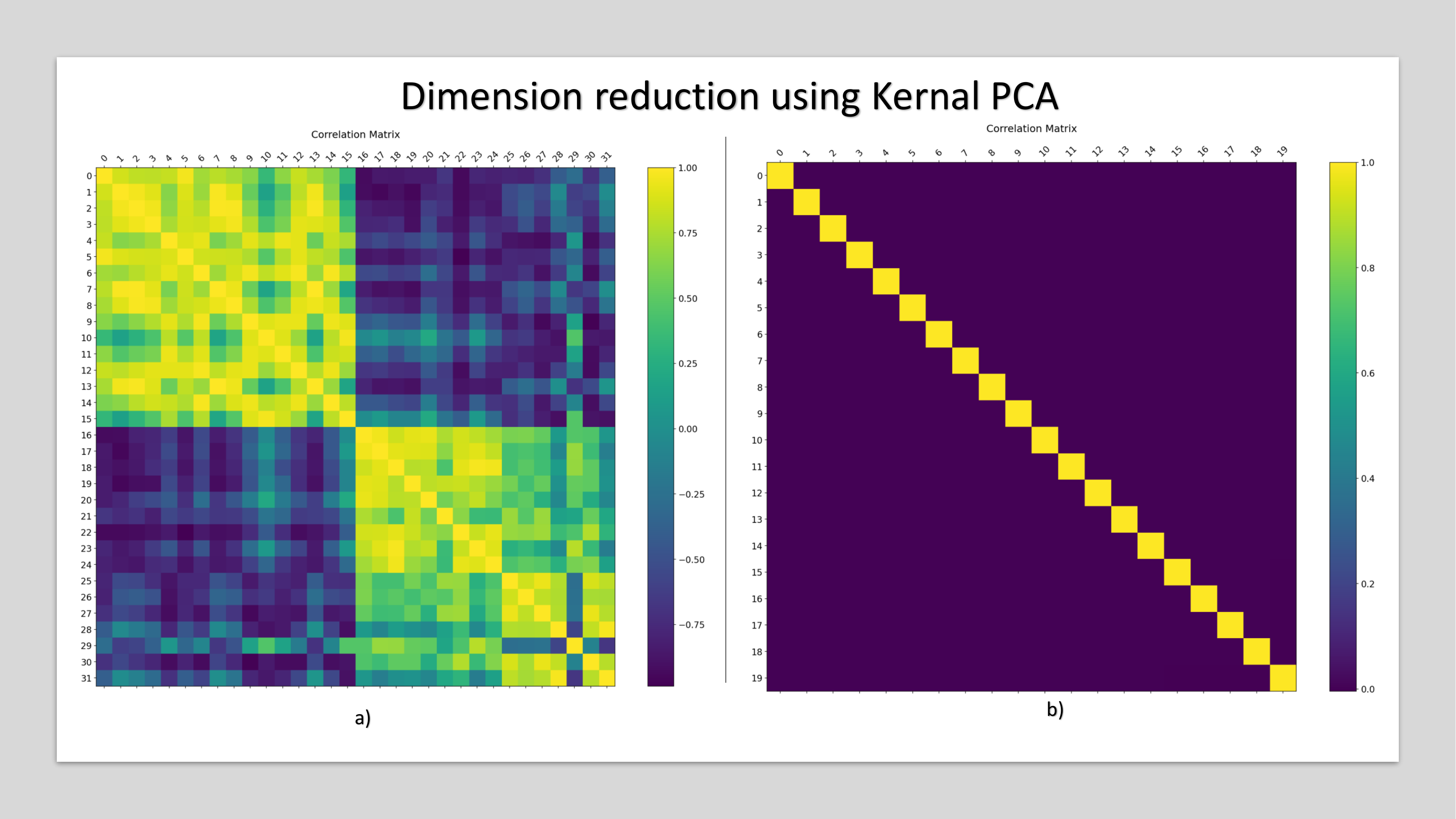}
        \caption{Correlation matrices a)Original data b) Kernal PCA with 20 components}
        \label{fig:corr_plot}
    \end{figure}

Some of the most common data streams, such as EEG and NIRS data, have high dimensionality. The collinearity with multiple variables is highly correlated. This condition may cause most of the machine learning algorithms to perform poorly. Hence it is essential to use methods such as Independent Component Analysis(ICA) or Principal Component Analysis(PCA) to reduce the dimensionality and maximize the statistical independence of the estimated components\cite{comon1994independent}. In this study, kernel PCA, an extension to the traditional PCA technique with the ability to extract principal nonlinear components without expensive computations, is used instead of PCA due to the nonlinear nature of the data\cite{mika1999kernel}. Fig.\ref{fig:corr_plot} illustrates how correlation matrices change after kernel PCA operation is applied to the original data to reduce dimensionality.

\begin{figure}[h]
        \centering
        \includegraphics[scale=0.3]{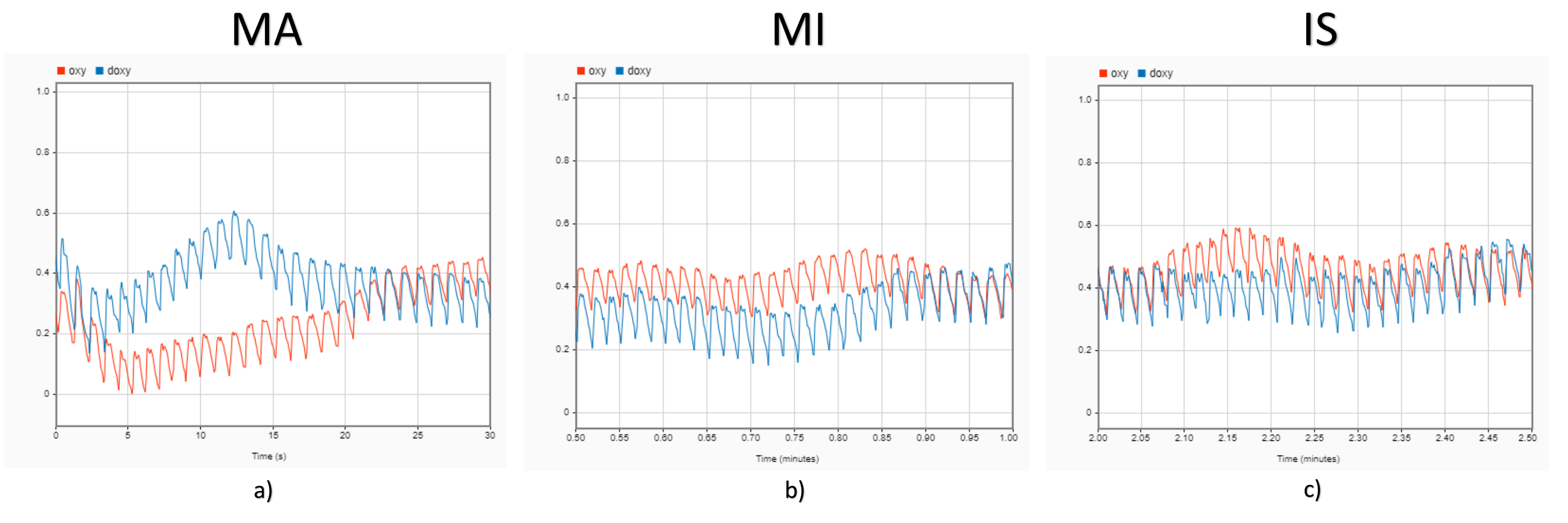}
        \caption{Comparison of the a)MA, b)MI and c)IS task in time domain with oxy and deoxy channels averaged together for subject 1 }
        \label{fig:task}
    \end{figure}

Another critical aspect of data visualization was the visual inspection of each of the time domain tasks. Fig.\ref{fig:task} shows how the oxy and deoxy channels were averaged for each task for the subject 1. The most important observation from this visualization is that there was a visually significant difference in the MA task than other tasks. The MI and IS tasks were visually indistinguishable in most cases. Later in our analysis, after training the models and obtaining the results, it was found that the classifiers have a higher ability to distinguish the MA task from the rest. 

\subsection{Pre-Processing}
The detected optical densities (ODs) of fNIRS data were converted to hemodynamic variations (concentration change in reduced hemoglobin HbR and concentration change in oxidized hemoglobin HbO) using the formula proposed by Matcher et al. .\cite{matcher1995performance}. The converted HbR and HbO values were band-pass filtered using a 3rd-order Butterworth zero-phase filter with a passband of 0.01–0.09 Hz to remove noise.

NIRS data were segmented into epochs from -5 to 25~s.  Baseline correction was performed by subtracting the temporal mean value between 1 and 0 s from each NIRS epoch. NIRS feature vectors were constructed using the temporal mean values of HbR and HbO in the 5-10~s and 10-15~s temporal windows in NIRS epochs from all channels, considering the inherent hemodynamic delay. 

\subsection{Feature Extraction and Selection}
The feature extraction and feature selection were made in several steps. For comparison, several different feature sets were extracted in the initial phase. A later dimensional reduction method was used to reduce the size of the data, which is crucial to eliminate overfitting. 

For the first feature, set a 2s moving window with a 50\% overlap to calculate each channel's mean,  peak, skewness, and kurtosis. The other feature set consisted of average power over the same window setting calculated over the frequency bands of [1-3]Hz and [4-6]Hz. The considered frequency bands were selected after considering several sub frequency bands in the range of [1-6]Hz.

\subsection{Model}
Learning from past information is a crucial part when analyzing time-series data such as fNIRS data. In principle,  recurrent networks (RNNs) can store data over extended time interval\cite{hochreiter1997long}. LSTM networks improve RNNs, which can learn long-term temporal dependencies by handling the vanishing gradient problem. The network can learn to bridge considerable time intervals while still keeping short time lag capabilities. LSTM  networks were used in translation, text prediction, natural language processing, audio, and image analysis. 

Bidirectional recurrent neural networks can learn from both past and future states, which can be beneficial when task classification is being done. Hence a Bi-Directional LSTM-based model is proposed for fNIRS based ternary task classification. This model was trained using raw data, which was dimensionally reduced to 20 components using ICA.

\begin{figure}[h]
        \centering
        \includegraphics[scale=0.30]{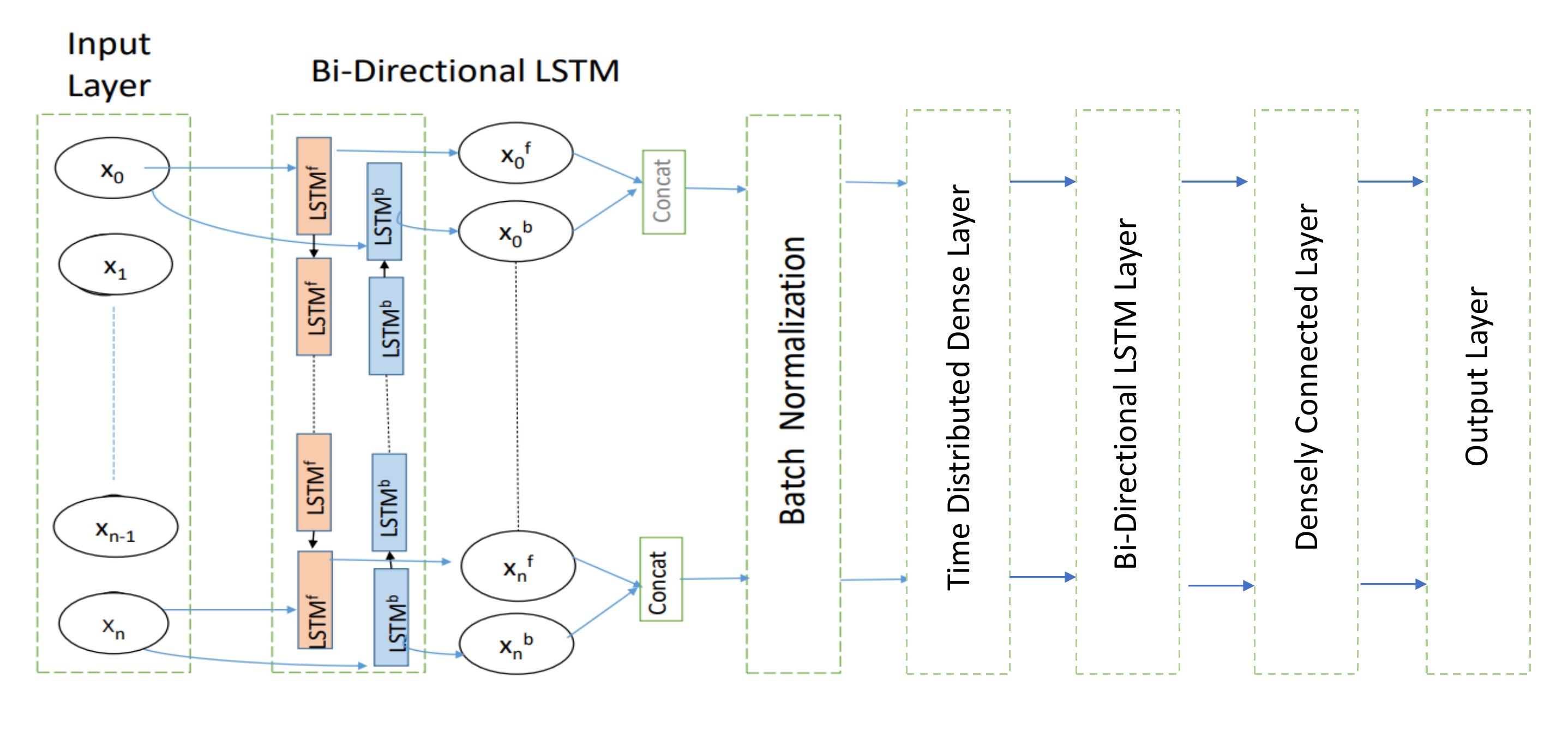}
        \caption{The proposed Bi-Directional LSTM based model for Task Classification }
        \label{fig:model}
    \end{figure}
    \begin{equation}\label{eq:relu}
    ReLU= max(0,x)
    \end{equation}
    
The model contains Bi-Directional LSTM layers and Densely Connected Layers, as illustrated in Fig.\ref{fig:model}. The input layer is a time distributed Dense layer, which takes a 3d input. The shape is defined according to the number of timesteps. The data from the input layer is fed into a bidirectional LSTM layer.  The recurrent function in Bi-Direction layers is activated by the Relu function, as shown in Eq.\ref{eq:relu}. The BiDirectional- LSTM layer is followed by a Batch Normalization layer, which feeds its output to another BiDirectional LSTM and Densely connected layer combination. 

The Time-Distributed Dense layers are activated by Scaled Exponential Linear Unit (SELU), as shown in Eq.\ref{eq:SELUfunc}. SELU function facilitates internal normalization, which is much faster than external normalization and helps the network converge faster. Further, the SELU function eliminates the vanishing and exploding gradient problem as well. In a Dense layer, each neuron receives input from all the neurons in the previous layer, hence densely connected.

 \begin{equation}\label{eq:SELUfunc}
   SELU(x)=\lambda \left\{\begin{matrix}
x & if x> 0\\ 
\alpha e^{x}-\alpha & if x\leq  0
\end{matrix}\right.
    \end{equation}

All the models used a softmax activation function for the output layer to obtain probabilities for classification.  Dropout is applied to the output of all hidden layers, and all layers have l2-kernel regularizer of strength 0.1 \cite{cogswell2015reducing}. The learning rate was reduced on the plateau, and early stopping was used to reduce overfitting. For all the models, the loss function used was categorical cross-entropy. The optimizer was  Nesterov accelerated Adaptive Moment Estimation (Nadam)\cite{dozat2016incorporating}.

   { \subsection{Training and Tuning Hyper-parameters}\label{subsec:hyperparameter}}
    
The proposed Bi-Directional LSTM architecture in this paper was trained, validated, and tested by splitting the dataset into three groups, train, validation, and test. The set of hyperparameters corresponding to the lowest classification error was selected after a grid search. The optimized hyperparameters include learning rate, dropout,  and the size of hidden units in LSTM layers using grid search. 
    
 An overfitted model may memorize the training data without learning to give good results on the training database. In this study, several schemes were used to prevent overfitting. One of the mechanisms was to use dropouts. A recurrent dropout was specified for LSTM layers, and 10\% for each of the recurrent layers.  Dropping out part of the system during the training phase prevents the neurons from adapting exceedingly well to the training data to reduce overfitting. As the neurons are dropped out, the connecting weights are excluded from updating and forcing the network to learn from the imperfect patterns and improve the model's generalization.
    
Early stopping was used when training was used to reduce overfitting for all the models\cite{neumaier1998solving}. Gaussian noise layer was also added during the training period to reduce overfitting for all the models. Further, for CNN and CNN+LSTM networks, other regularization methods such as batch normalization were used. The batch normalization layer normalizes a previous activation layer's output by subtracting the batch mean and dividing by the batch standard deviation. Further, LeCun normal initializer is used for weight initialization of LSTM networks. 
    
For all the classifiers, the entire data set was split into 70:30 train and test set. Then from the training set, a 70:30 split was used for train and validation. The implementation of all three models was done with python libraries Keras and TensorFlow. Each of the models was trained for 100 epochs with early stopping when the loss is no longer improving using four input batches.
\section{Results}
In this study, the indicators used for testing are classification accuracy and the area under the curve. The definitions of the above indicators are as follows:

\begin{equation}\label{Accuracy}
\mathit{Accuracy = \dfrac{TP + TN}{TP + FP + TN + FN}}
\end{equation}

Where TP is the number of true positives, FN is the number of false negatives, FP is the number of false
positives and TN is the number of true negatives. 

\begin{figure}[h]
        \centering
        \includegraphics[scale=1]{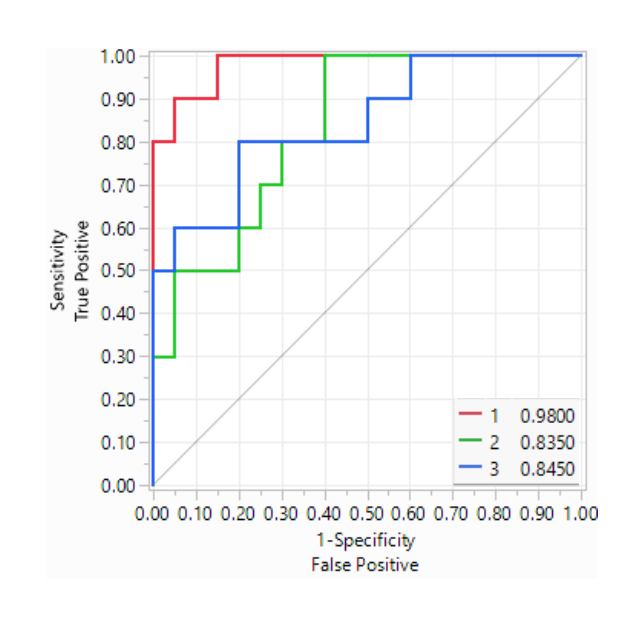}
        \caption{Receiver Operator Characteristic for subject 1 obtained using LSTM model}
        \label{fig:roc_sub1l}
    \end{figure}

ROC curves typically feature a true positive rate on the Y-axis and false positive rate on the X-axis. A more substantial area under the curve (AUC) is usually considered better. AUC is an important metric for the classifier's ability to distinguish between the classes accurately. An AUC value over 0.9 considers signifying the first classifier, while over 0.8 can be considered a good classifier. The proposed model has different AUCs for different tasks. The AUC values vary across the subjects as well. Fig.~ \ref{fig:roc_sub1l} shows the ROC curve for the LSTM model trained with subject 1 data. Tasks 1,2, and 3 correspond to the MA, MI, and IS tasks, respectively. From the figure, it is demonstrated that the differentiation between the MA task and the others is much better than differentiating between the other two graphs. The area under the ROC curve is 0.98, for the MA task, which shows an excellent ability to differentiate from the rest.  The classifier can be considered excellent in differentiating between the two classes considering the AUC value. 

The original data classifier used by Shin et al. used a ten × 10-fold cross-validation with shrinkage linear discriminant analysis (sLDA) for each of the three binary classification problems. The sLDA can effectively minimize the negative effect resulting from high-dimensional feature vectors compared to the number of trials. This sLDA based system obtained an average classification accuracy of 64.1\%.

     \begin{figure}[h]
        \centering
        \includegraphics[scale=0.6]{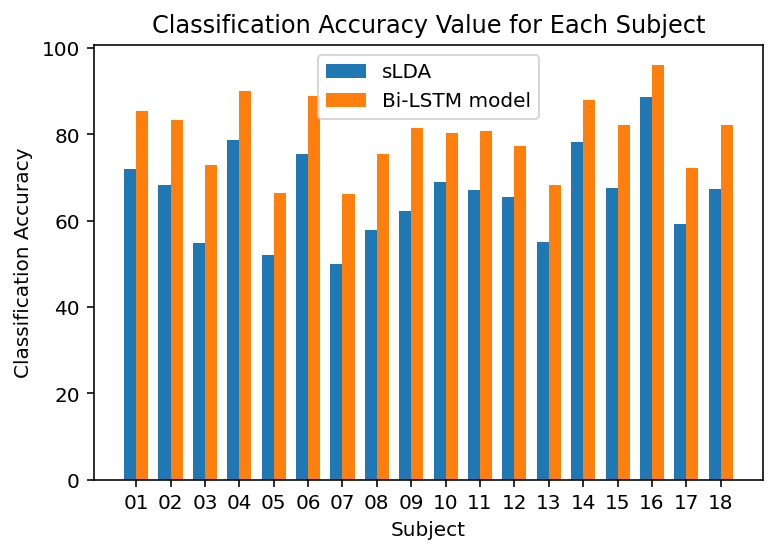}
        \caption{Comparison of Classification Accuracy between the sLDA and proposed Bi-Directional LSTM model for all the 18 subjects using the same feature set}
        \label{fig:clas_acc}
    \end{figure}
    
    \begin{figure}[h]
        \centering
        \includegraphics[scale=0.6]{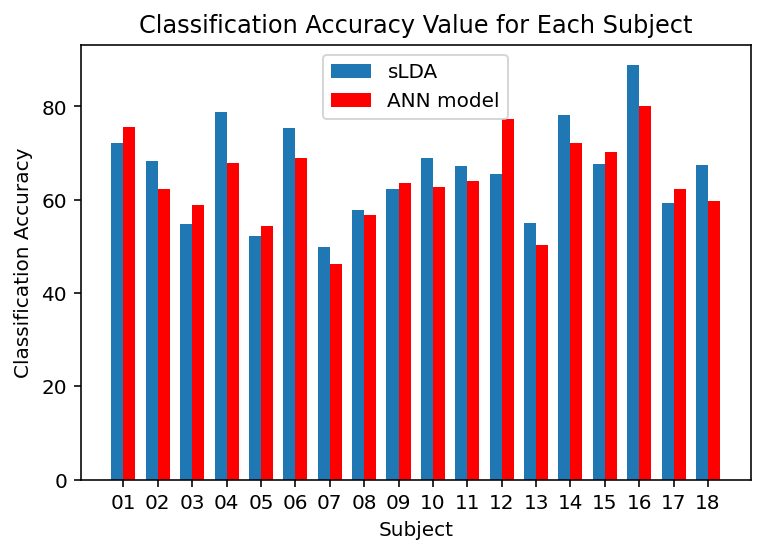}
        \caption{Comparison of Classification Accuracy between the sLDA and ANN model for all the 18 subjects using the same feature set}
        \label{fig:clas_acc}
    \end{figure}

The confusion matrix for the proposed LSTM classifier is given in Table.\ref{tab:model_results}. A total of 14 test samples were used for the testing phase. The classifier predicted whether the patient previously suffered a concussion or not. The accuracy of the proposed model is 81.48\% based on the test data. 
\begin{table}[]
\caption{Confusion Matrix for Proposed Bi-Directional LSTM model using test data from subject 1}
\label{tab:model_results}
\centering
\begin{tabular}{|l|l|l|l|}
\hline
\multirow{2}{*}{\textbf{Actual}} & \multicolumn{3}{l|}{\textbf{Predicted}} \\ \cline{2-4} 
                                 & MA          & MI          & IS          \\ \hline
MA                               & 9           & 0           & 1           \\ \hline
MI                               & 1           & 7           & 2           \\ \hline
IS                               & 0          & 1           & 9           \\ \hline
\end{tabular}
\end{table}

\begin{table}[]
\caption{Confusion Matrix for ANN  model using test data from subject 1}
\label{tab:ANN_conf}
\centering
\begin{tabular}{|l|l|l|l|}
\hline
\multirow{2}{*}{\textbf{Actual}} & \multicolumn{3}{l|}{\textbf{Predicted}} \\ \cline{2-4} 
                                 & MA          & MI          & IS          \\ \hline
MA                               & 8           & 0           & 2           \\ \hline
MI                               & 2           & 6          & 2           \\ \hline
IS                               & 1           & 2           & 7           \\ \hline
\end{tabular}
\end{table}
\subsection{Baseline Performance}
Before training the deep learning-based classifier, the traditional machine learning classifiers such as logistic regression, SVM, and simple Artificial Neural Networks were used to establish the baseline performance. The traditional classifiers are popular due to the simplicity. These classifiers are usually easy to implement and requires fewer computations than deep-learning classifiers. The classifiers were tested separately with features extracted from raw data and dimension reduced data. One disadvantage of these baseline classifiers is the extensive pre-processing that has to be performed before training the models. Hence even though the classification method may be simple, the pre-processing step can be computationally expensive to handle. From all the conventional machine learning methods, the artificial neural network system had the best accuracy with 64.5\%. The Fig.\ref{fig:clas_acc} shows the accuracy comparison between the ANN model and the sLDA classifier proposed by Shin et.al\cite{shin2018ternary}. The ANN classifier obtained a slightly higher accuracy than the sLDA classifier and used the same feature set. The confusion matrix for the ANN classifier with test data for subject one is given in Table\ref{tab:ANN_conf}. As seen from the confusion matrix, the MA task has the least number of misclassifications, and this condition remains true for other subjects as well.
\subsection{Deep Learning Model Performance}
There were several deep learning models considered initially to be trained by different sets of features and levels of pre-processing. The model with the best performance parameters, which is classification accuracy, was chosen as the proposed model from these models. The proposed model is a Bi-Directional LSTM model trained using raw data with dimensions reduced to 20 using the ICA algorithm. The average classification accuracy obtained was 81.48\%. The accuracy comparison between the proposed classifier and the sLDA classifier is given in Fig.\ref{fig:clas_acc}. The confusion matrix for the proposed classifier using subject 1 test data is given in Table \ref{tab:model_results}. The ROC curves for the proposed system are shown in Fig.\ref{fig:roc_sub1l} obtained using subject 1 test data. As seen from the figure, the AUROC for the MA task is much higher than the other tasks.

\section{Discussion}
The results of our study show that NIRS based classification of tasks can be increased by using deep learning methods. This can be an essential step to enhance NIRS based classification systems used for BCI applications. The proposed ternary classification system can classify MA, MI, and IS-related brain activation patterns.  The deep learning models often suffer from overfitting due to high dimensional data. This was one of the most significant problems encountered in this study.  Several measures were taken to overcome this problem using dimension reduction techniques, introducing Gaussian noise while training the model, and regularization techniques. By increasing these measures, we increase the computational burden as well. Yet the classification results are overall poor when raw data was used for classification tasks. 

From the classification results across the subject, especially by analyzing the ROC curves, the most critical observation is that the AUROC value for the MA task is considerably higher than the other two curves. This fact holds when analyzing the confusion matrix and fewer misclassifications compared to the other two categories. The classifiers are overall much better in detecting the MA task than the other two tasks. This observation holds for all subjects. The reason for this phenomenon may be because we are using the fNIRS data acquired from the front region of the frontal lobe. The MI task is associated with activating the mid-region of the frontal lobe. The data was not collected from that area due to practical difficulties with the setup. Some valuable information is lost by not being able to acquire data from that region. Hence classification systems that use purely fNIRS data may not perform well with classifying activities such as Motor Imagery.  In the original publication, which used this data set by Shin et al ., EEG data overcame this problem. Using a hybrid system, different brain activated areas during various tasks can be captured through the multi-modal approach\cite{shin2018ternary}.

A hybrid BCI application with a deep learning classification system has many future potentials with increased computation capabilities. One of the most important aspects of deep learning-based models is handling the raw data without much pre-processing. These models can find the features by themselves rather than following the traditional feature extraction and feature selection approaches. By using raw data, more complex deep learning architectures may be required for classification tasks. With the complexity of the classifier will require more computational resources will also be required. In this particular instance, the raw data based classifier performed poorly than the others. This result does not mean that future classifiers should abandon raw data based classifiers. Much better tuning and novel architectures can be a solution to improve the performance of such classifiers.

\section{Conclusion}
This paper presented a deep learning classifier using fNIRS data. fNIRS data is increasingly popular with research on BCI interfaces. Traditionally, simpler classification systems such as SVM or LDR were used for classification tasks. With the availability of higher computational power, deep learning-based methods are increasingly used as classification systems. Deep Learning systems improve accuracy when the training data follows complicated patterns, usually the case in BCI based applications. The method proposed has an accuracy of  81.48\%. The proposed method is an improvement over the previous ternary classification system. The deep learning model can be improved by integrating other modalities like EEG data to detect brain signals. One drawback of using deep learning is the black-box approach, which may not give a useful description on which channels are more prominent in predicting the task. With explainable deep learning gaining momentum, the models will also be explainable to an extent in the future. For further studies, one significant development will be modifying the deep learning-based system to be trained using the raw data, eliminating the complicated pre-processing process associated with the fNIRS data.

\bibliographystyle{IEEEtran}
\bibliography{fnirs}

\begin{thebibliography}{10}
\providecommand{\url}[1]{#1}
\csname url@samestyle\endcsname
\providecommand{\newblock}{\relax}
\providecommand{\bibinfo}[2]{#2}
\providecommand{\BIBentrySTDinterwordspacing}{\spaceskip=0pt\relax}
\providecommand{\BIBentryALTinterwordstretchfactor}{4}
\providecommand{\BIBentryALTinterwordspacing}{\spaceskip=\fontdimen2\font plus
\BIBentryALTinterwordstretchfactor\fontdimen3\font minus
  \fontdimen4\font\relax}
\providecommand{\BIBforeignlanguage}[2]{{%
\expandafter\ifx\csname l@#1\endcsname\relax
\typeout{** WARNING: IEEEtran.bst: No hyphenation pattern has been}%
\typeout{** loaded for the language `#1'. Using the pattern for}%
\typeout{** the default language instead.}%
\else
\language=\csname l@#1\endcsname
\fi
#2}}
\providecommand{\BIBdecl}{\relax}
\BIBdecl

\bibitem{naseer2015fnirs}
N.~Naseer and K.-S. Hong, ``fnirs-based brain-computer interfaces: a review,''
  \emph{Frontiers in human neuroscience}, vol.~9, p.~3, 2015.

\bibitem{sassaroli2004comment}
A.~Sassaroli and S.~Fantini, ``Comment on the modified beer--lambert law for
  scattering media,'' \emph{Physics in Medicine \& Biology}, vol.~49, no.~14,
  p. N255, 2004.

\bibitem{blankertz2016berlin}
B.~Blankertz, L.~Acqualagna, S.~D{\"a}hne, S.~Haufe, M.~Schultze-Kraft,
  I.~Sturm, M.~U{\v{s}}{\'c}umlic, M.~A. Wenzel, G.~Curio, and K.-R.
  M{\"u}ller, ``The berlin brain-computer interface: progress beyond
  communication and control,'' \emph{Frontiers in Neuroscience}, vol.~10, p.
  530, 2016.

\bibitem{abdalmalak2020assessing}
A.~Abdalmalak, D.~Milej, L.~Yip, A.~R. Khan, M.~Diop, A.~M. Owen, and
  K.~St~Lawrence, ``Assessing time-resolved fnirs for brain-computer interface
  applications of mental communication,'' \emph{Frontiers in Neuroscience},
  vol.~14, p. 105, 2020.

\bibitem{abiri2019comprehensive}
R.~Abiri, S.~Borhani, E.~W. Sellers, Y.~Jiang, and X.~Zhao, ``A comprehensive
  review of eeg-based brain--computer interface paradigms,'' \emph{Journal of
  neural engineering}, vol.~16, no.~1, p. 011001, 2019.

\bibitem{shin2018ternary}
J.~Shin, J.~Kwon, and C.-H. Im, ``A ternary hybrid eeg-nirs brain-computer
  interface for the classification of brain activation patterns during mental
  arithmetic, motor imagery, and idle state,'' \emph{Frontiers in
  neuroinformatics}, vol.~12, p.~5, 2018.

\bibitem{hong2017hybrid}
K.-S. Hong and M.~J. Khan, ``Hybrid brain--computer interface techniques for
  improved classification accuracy and increased number of commands: a
  review,'' \emph{Frontiers in neurorobotics}, vol.~11, p.~35, 2017.

\bibitem{li2019advances}
Z.~Li, S.~Zhang, and J.~Pan, ``Advances in hybrid brain-computer interfaces:
  Principles, design, and applications,'' \emph{Computational Intelligence and
  Neuroscience}, vol. 2019, 2019.

\bibitem{tanveer2019enhanced}
M.~A. Tanveer, M.~J. Khan, M.~J. Qureshi, N.~Naseer, and K.-S. Hong, ``Enhanced
  drowsiness detection using deep learning: An fnirs study,'' \emph{IEEE
  Access}, vol.~7, pp. 137\,920--137\,929, 2019.

\bibitem{chiarelli2018deep}
A.~M. Chiarelli, P.~Croce, A.~Merla, and F.~Zappasodi, ``Deep learning for
  hybrid eeg-fnirs brain--computer interface: application to motor imagery
  classification,'' \emph{Journal of neural engineering}, vol.~15, no.~3, p.
  036028, 2018.

\bibitem{comon1994independent}
P.~Comon, ``Independent component analysis, a new concept?'' \emph{Signal
  processing}, vol.~36, no.~3, pp. 287--314, 1994.

\bibitem{mika1999kernel}
S.~Mika, B.~Sch{\"o}lkopf, A.~J. Smola, K.-R. M{\"u}ller, M.~Scholz, and
  G.~R{\"a}tsch, ``Kernel pca and de-noising in feature spaces,'' in
  \emph{Advances in neural information processing systems}, 1999, pp. 536--542.

\bibitem{matcher1995performance}
S.~Matcher, C.~Elwell, C.~Cooper, M.~Cope, and D.~Delpy, ``Performance
  comparison of several published tissue near-infrared spectroscopy
  algorithms,'' \emph{Analytical biochemistry}, vol. 227, no.~1, pp. 54--68,
  1995.

\bibitem{hochreiter1997long}
S.~Hochreiter and J.~Schmidhuber, ``Long short-term memory,'' \emph{Neural
  computation}, vol.~9, no.~8, pp. 1735--1780, 1997.

\bibitem{cogswell2015reducing}
M.~Cogswell, F.~Ahmed, R.~Girshick, L.~Zitnick, and D.~Batra, ``Reducing
  overfitting in deep networks by decorrelating representations,'' \emph{arXiv
  preprint arXiv:1511.06068}, 2015.

\bibitem{dozat2016incorporating}
T.~Dozat, ``Incorporating nesterov momentum into adam,'' 2016.

\bibitem{neumaier1998solving}
A.~Neumaier, ``Solving ill-conditioned and singular linear systems: A tutorial
  on regularization,'' \emph{SIAM review}, vol.~40, no.~3, pp. 636--666, 1998.

\end{thebibliography}

\end{document}